\documentclass{article}
\usepackage{ijcai23}

\usepackage{times}
\usepackage{soul}
\usepackage{url}
\usepackage[hidelinks]{hyperref}
\usepackage[utf8]{inputenc}
\usepackage[small]{caption}
\usepackage{graphicx}
\usepackage{amsmath}
\usepackage{amsthm}
\usepackage{booktabs}
\usepackage{algorithm}
\usepackage{algorithmic}
\usepackage[switch]{lineno}

\newif\ifcomments
\commentstrue
\ifcomments
    \usepackage{xcolor}
    \newcommand\rob[1]{\textcolor{violet}{[RV: #1]}}
\else
    \newcommand\rob[1]{}    
\fi

\title{GreenPLM: Cross-Lingual Transfer of Monolingual Pre-Trained Language Models at Almost No Cost}

\author{
Qingcheng Zeng$^{1,2}$\thanks{Equal contribution.}\and
Lucas Garay$^{1\ast}$\and
Peilin Zhou$^{1,3\ast}$\and
Dading Chong$^4$\and
Yining Hua$^5$\and
Jiageng Wu$^1$\and
Yikang Pan$^1$\and
Han Zhou$^6$\and
Rob Voigt$^2$\and
Jie Yang$^1$\thanks{\ Corresponding author.}
\affiliations
$^1$School of Public Health and the Second Affiliated Hospital, Zhejiang University\\
$^2$Department of Linguistics, Northwestern University\\
$^3$Data Science and Analytics Thrust, Hong Kong University of Science and Technology (Guangzhou)\\
$^4$School of Electronic and Computer Engineering, Peking University\\
$^5$Department of Biomedical Informatics, Harvard University\\
$^6$Language Technology Lab, University of Cambridge
\emails
jieynlp@gmail.com}

\begin{document}

\maketitle

\begin{abstract}
Large pre-trained models have revolutionized natural language processing (NLP) research and applications, but high training costs and limited data resources have prevented their benefits from being shared equally amongst speakers of all the world's languages. To address issues of cross-linguistic access to such models and reduce energy consumption for sustainability during large-scale model training, this study proposes an effective and energy-efficient framework called GreenPLM that uses bilingual lexicons to directly ``translate'' pre-trained language models of one language into another at almost no additional cost. We validate this approach in 18 languages' BERT models and show that this framework is comparable to, if not better than, other heuristics with high training costs. In addition, given lightweight continued pre-training on limited data where available, this framework outperforms the original monolingual language models in six out of seven tested languages with up to 200x less pre-training efforts. Aiming at the Leave No One Behind Principle (LNOB), our approach manages to reduce inequalities between languages and energy consumption greatly. We make our codes and models publicly available here: \url{https://github.com/qcznlp/GreenPLMs}
\end{abstract}

\section{Introduction}\label{sec1}
In recent years, NLP has welcomed great advances and significant progress in deep learning \cite{vaswani2017attention,devlin-etal-2019-bert,NEURIPS2020_1457c0d6}. Pre-trained language models (PLMs) trained on extensive text have significantly improved performance on various core NLP tasks, bringing NLP new research paradigms such as ``pre-training and fine-tuning" and ``prompt-based learning" \cite{10.1145/3560815}. However, despite PLMs' extraordinary performance, training these models both incurs substantial computational costs and relies upon the availability of extensive training data. The former has raised public concern about substantial energy consumption and environmental damage \cite{Strubell_Ganesh_McCallum_2020}, while the latter inevitably creates unequal opportunities for NLP research across languages with differing access to resources \cite{joshi-etal-2020-state}.

\begin{figure*}[t]
  \centering
  \includegraphics[width=\textwidth]{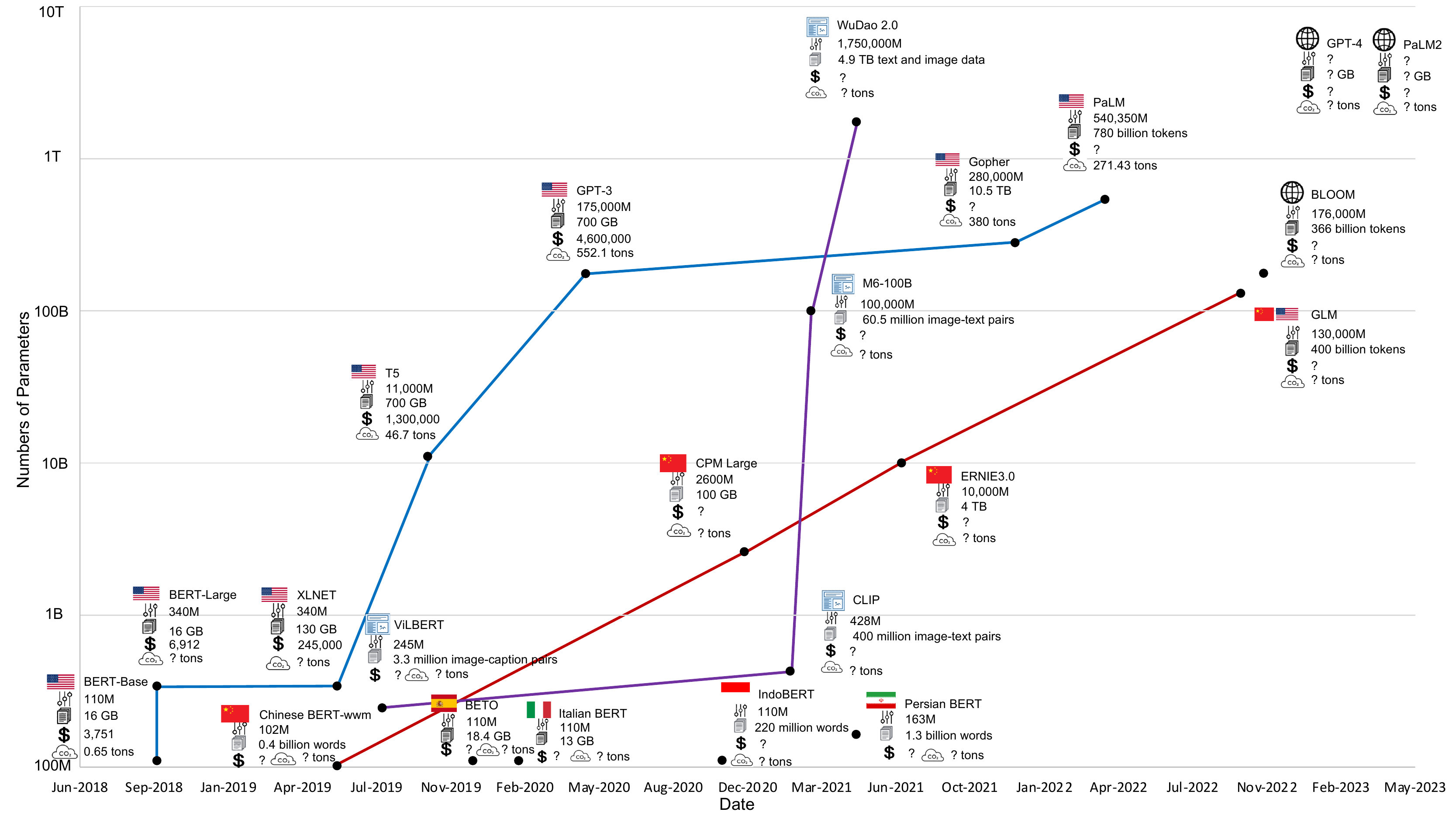}
  \caption{Trends of Complexity and Training Cost in Mainstream PLMs with Significant Performance Boosts. The icons represent the number of parameters, training corpus size, financial cost by dollars, and the emission of CO\begin{math}_{2}\end{math}, respectively. ? represents numbers that can not be estimated.}
\label{fig:trend}
\end{figure*}

Computation costs for training a single, state-of-the-art deep learning model expanded 300,000 times between 2012 and 2018, a trend which has only intensified since; this presents a challenge for researchers interested in upholding green AI principles \cite{https://doi.org/10.48550/arxiv.1907.10597} as well as broader public policy ambitions like the United Nations Sustainable Development Goals (SDG) \cite{sdg30}. Figure~\ref{fig:trend} shows a comparison of a number of representative PLMs and multimodal models from BERT-base \cite{peters-etal-2018-deep} to GPT-4 \cite{openai2023gpt4}, where we observe trends of ever-increasing model size, computational costs, economic costs, and environmental costs \cite{strubell-etal-2019-energy}. 
Pre-training a BERT-base model incurs cloud computing costs of \$3,751 to \$12,571 \cite{https://doi.org/10.48550/arxiv.1907.10597}, which is unaffordable for personal use and most researchers in developing countries. Training models such as BERT, T5 \cite{JMLR:v21:20-074}, PaLM, Gopher \cite{https://doi.org/10.48550/arxiv.2112.11446}, and GPT-3 \cite{NEURIPS2020_1457c0d6} from scratch emits 0.65, 46.7, 271.43, 380, and 552.1 tons of CO\begin{math}_{2}\end{math}, respectively. 
 
Beyond economic and environmental costs, the pre-training of large language models requires vast amounts of monolingual text data, which is lacking in many low-resource languages, leading to increased disparities in the utilization of language models among different languages. Languages with a larger number of speakers inherently have larger corpora available for PLM training. This leaves only a tiny fraction of the approximately 7000 languages worldwide with sufficient data to train a monolingual PLM \cite{ebrahimi-kann-2021-adapt}. Table~\ref{tab:1} lists the sizes of training materials for the monolingual BERT of several languages, exhibiting a significant inequality in monolingual materials. For instance, English GPT-3 was trained on 45 TB of monolingual material, equivalent to 4,608 times size of the Hindi BERT pre-training corpus. This basic inequality in data availability is further exacerbated by many social and financial factors, such as internet infrastructure, education, and computational resources available in low-resource speech communities. Self-supervised pre-training thus enlarges the disparity between high- and low-resource languages, with the result that breakthrough technologies in contemporary NLP have proven particularly difficult to share equally among speakers of all the world's languages.

\begin{figure}[]
  \centering
  \includegraphics[width=\columnwidth]{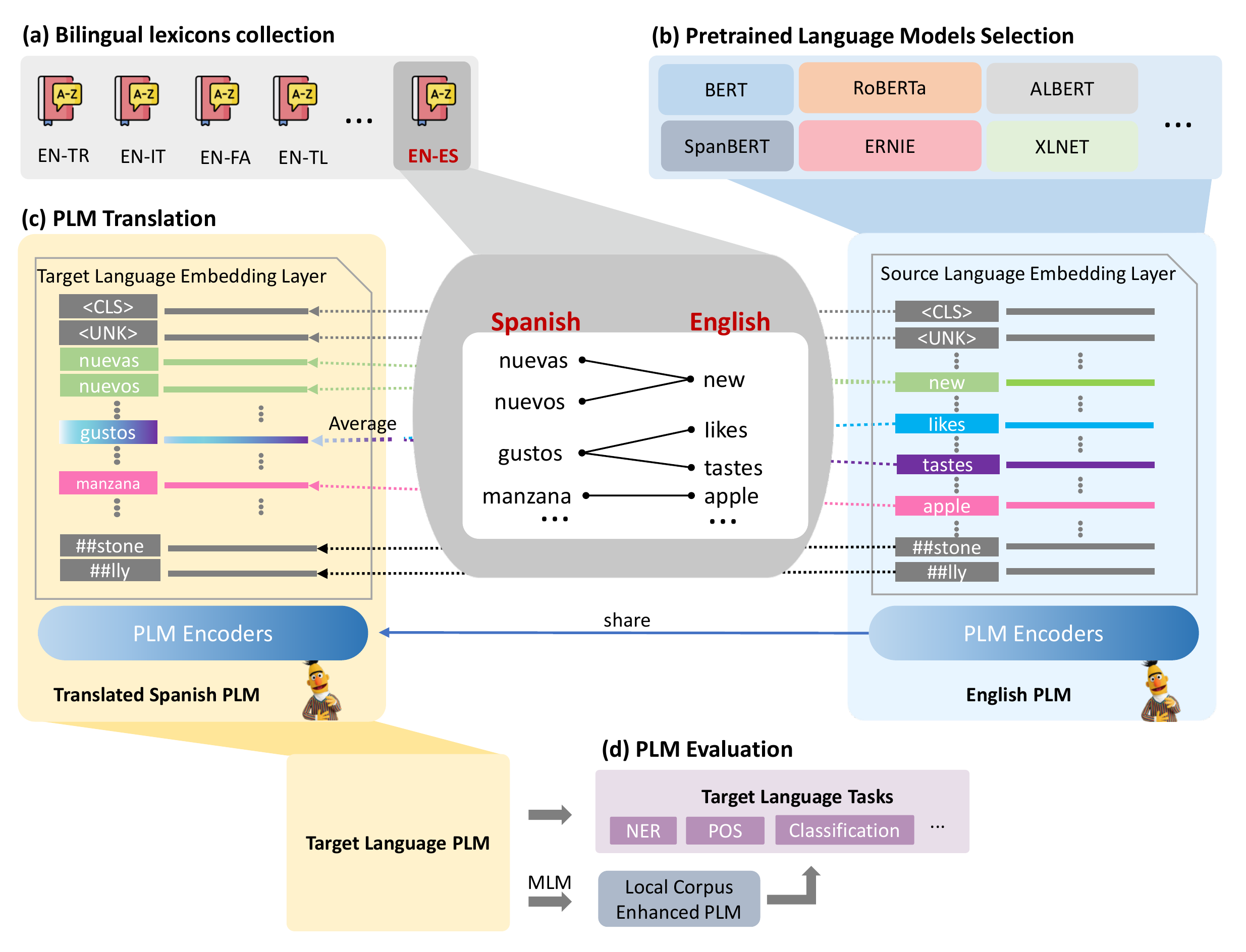}
  \caption{Overview of the GreenPLM approach. Bilingual lexicons (a) are used to augment the vocabulary and embedding layer of a source PLM (b) using our proposed Lexicon Walk Mapping strategy (c), while holding the parameters and architecture of the source model constant, resulting in a translated model that can be further fine-tuned towards particular target-language tasks or continually pre-trained using a smaller dataset in the translated language (d).}
\label{fig:method}
\end{figure}

To contribute to ongoing efforts to mitigate these issues, we propose a generalizable framework, GreenPLM, to ``translate'' monolingual PLMs to new languages at almost no additional cost. We hypothesize that the linguistic knowledge learned by PLMs on large monolingual corpora is transferable to low-resource languages, encouraged by two primary pieces of evidence. First, monolingual word embeddings are partially isomorphic across languages \cite{artetxe-etal-2016-learning}; etymologically close language pairs show good word translation performance \cite{conneau2017word}. Second, work 
on the capacities of monolingual PLMs \cite{conneau-etal-2020-emerging,blevins-zettlemoyer-2022-language} has shown that they encode cross-linguistically useful information even without training in further languages. Therefore, GreenPLM utilizes bilingual lexicons to bridge cross-lingual semantic spaces while keeping the well-trained parameters of high-resource PLMs fixed as shown in Figure \ref{fig:method}. Our experimental results show that GreenPLM's performance is comparable to or better than current heuristics in 18 tested languages, at nearly zero cost. In addition, with around 0.5\% computational cost compared to pre-training a BERT model from scratch, continued pre-training of these models (which we call GreenPLM+) outperforms the performance of monolingual PLMs. To sum up, our contributions to AI and social good are the following:
\begin{itemize}
    \item [1)] We propose a simple, heuristic pipeline utilizing bilingual lexicons to translate a source language PLM to a target language PLM with almost no computational cost, significantly reducing carbon emissions for building foundation PLMs in various languages;
    \item [2)] We verify the effectiveness and generalizability of this framework with evaluations in 18 languages, showing performance comparable to monolingual and multilingual PLMs;
    \item [3)] We propose these models can be further enhanced with continued pre-training in 7 languages and show performance exceeding existing monolingual and multilingual PLMs while reducing pre-training costs up to 200x compared to pre-training from scratch;
    \item [4)] We suggest this approach can be used to reduce language disparities and energy consumption in pre-training LMs, aligned with the SDG and the LNOB principle to promote the application of AI for social good.
\end{itemize}

\section{Related Work}\label{sec2}
\subsection{Cost Reduction in PLM Training} 
As we've explored, training large PLMs inevitably requires extensive computational resources, which attracted many studies on efficient PLM pre-training. For example, \cite{https://doi.org/10.48550/arxiv.2006.03654} introduced a disentangled attention mechanism to separate token and position embeddings, reducing over 60\% of the costs of pre-training from scratch with an optimized implementation; \cite{https://doi.org/10.48550/arxiv.2111.04130} proposed a task-driven language modeling framework to pre-train task-specific PLMs that required only around 2\% of the original computational costs. An alternative approach explored by \cite{chen-etal-2022-bert2bert} and \cite{qin-etal-2022-knowledge}  leveraged a small ``teacher PLM" to pre-train a large ``student PLM" at a low computational cost. Reducing data requirements remains less studied compared to the former. The issue of data requirements has been less studied, though recently \cite{https://doi.org/10.48550/arxiv.2301.11796} put forth a BabyLM challenge to call for sample-efficient pre-training. Our work addresses both these issues, focusing on simultaneously reducing computational costs and data requirements.

\subsection{Language Disparity in PLMs}
Researchers have proposed numerous mechanisms to equalize performance across human languages in PLMs \cite{muller-etal-2021-unseen,ebrahimi-kann-2021-adapt}, but the democratization of language technology remains a major challenge \cite{bird-2020-decolonising}. Multilingual BERT (mBERT) represents a preliminary attempt to adapt PLMs to multiple languages and is pre-trained on the concatenation of corpora from different languages \cite{devlin-etal-2019-bert}. Such models can achieve substantial coverage; for example, mBERT and XLM-R \cite{conneau-etal-2020-unsupervised} cover 104 and 100 languages respectively. Nevertheless, model performances in specific languages are reported to be highly dependent on pre-training corpus sizes, and cross-lingual PLMs trained in this manner are still expensive to train and ultimately rely upon the availability of monolingual texts in a given target language.

\subsection{Cross-Lingual Transfer}
After multilingual BERT showed impressive performance in zero-shot cross-language transfer settings \cite{pires-etal-2019-multilingual}, substantial attempts have been made to adapt monolingual or multilingual PLMs to other languages. 

\cite{wu-dredze-2020-languages} examined the performance of mBERT across various languages and reported that the model performs well for high-resource languages but not for low-resource languages. Nonetheless, in low-resource settings, mBERT performs better than monolingual BERT, validating some degree of cross-language generalizability. \cite{muller-etal-2021-unseen} explored mBERT’s performance in never-before-seen languages, finding unstable performance. They found that mBERT has unstable performance across languages, depending on the training data. \cite{ebrahimi-kann-2021-adapt} used the New Testament, which is available in most languages, to adapt multilingual PLMs to more than 1600 languages using continued pre-training, vocabulary extension, and adapter transformers. \cite{https://doi.org/10.48550/arxiv.2203.09435} explored how bilingual lexicons can enhance the multilingual language coverage of PLMs. Their results had consistent improvement across sequence labeling and dependency parsing tasks in 19 low-resource languages. In contrast, our current study extends bilingual lexicons to monolingual settings as a bridge rather than data augmentation components to design specified monolingual PLMs equally for all languages.

Monolingual adaptation remains relatively under-explored compared to multilingual. It follows two general directions: extending monolingual PLMs to bilingual and adapting a source language PLM to a target language PLM. In the first direction, \cite{artetxe-etal-2020-cross} continually pre-trained English PLMs on data in target languages with the transformer body frozen and fine-tuned the PLMs on labeled English data. Their results confirmed the generalizability of monolingual PLMs across languages. \cite{https://doi.org/10.48550/arxiv.2002.07306} transferred PLMs by initializing foreign language embeddings in the English vector space and jointly fine-tuning both models. For the second direction, \cite{de-vries-nissim-2021-good}, and \cite{https://doi.org/10.48550/arxiv.2112.06598} used word embeddings as the bridge to adapt monolingual models to other languages with continued pre-training. Motivated by these works, we leverage more widely available and less computationally expensive resources other than word embeddings to generalize PLMs.

\section{Methodology}
\subsection{Overall Architecture}
An overview of the GreenPLM framework is shown in Figure \ref{fig:method}. We start from a bilingual lexicon for a given source-target language pair and the selection of a source PLM. Importantly, contemporary PLMs require a pre-established vocabulary embedding layer which operates on the outputs of a specific tokenizer. Tokens in the vocabulary are a mix of both full words and word pieces, traditionally the outputs of an algorithm like WordPiece or byte pair encoding, where each token is mapped to an embedding learned in the pre-training process.

In our approach, we augment the vocabulary and embedding layer of the source PLM to incorporate target-language vocabulary from the bilingual lexicon, using a method to map this lexicon into a shared semantic space explained below in subsection \ref{lexiconwalkmapping}. This results in what we call a ``translated PLM,'' though notably the embedding layer still contains the entire vocabulary from the source language and the PLM encoder parameters are yet unchanged from the source PLM. Such a model can already be applied directly to target-language data; in experiments, we call this setting simply ``GreenPLM".

The translated PLM can also be fine-tuned towards particular tasks in the target language, or pre-trained further on target language data, both of which modify the PLM parameters to be more suited to the characteristics of the target language. We name this setting ``GreenPLM+''. 

\subsection{Lexicon Walk Mapping}\label{lexiconwalkmapping}
The biggest challenge in the approach outlined above is how to use a bilingual lexicon to generate coherent target-language embeddings that can be fed into source-language PLM encoder parameters. To achieve this we propose a heuristic method we call Lexicon Walk Mapping (LWM), in which a bilingual lexicon is used as the bridge to create a mapping between the semantic spaces of the source and target languages. 

We define \textbf{a bilingual lexicon} as \emph{L} = \begin{math}\{x_i, y_i\} \end{math} where \begin{math} x_i \end{math} is a word or phrase in the source language (English, in this work) and \begin{math} y_i \end{math} is \begin{math} x_i \end{math}'s translation in the target language. Then given a \textbf{source language's PLM \emph{S}} with its vocabulary \begin{math} S_{vocab} \end{math} and token embeddings \begin{math} S_{embedding} \end{math}, we walk through the lexicon and operate on each entry separately. For each entry, we first tokenize the source word or phrase \begin{math} x_i \end{math} into a set of tokens \emph{T} = \begin{math} \{t_1, t_2, ...t_i...t_n \mid t_i \in S_{vocab} \end{math}\begin{math}\} \end{math}, and then map \emph{T} to the token embedding layer of \emph{S} to retrieve a set of embeddings \emph{E} = \begin{math} \{e_1, e_2, ...e_i...e_n \mid e_i \in S_{embedding} \end{math}\begin{math}\} \end{math}. We then take the dimension-wise average of \emph{E}, and the new embedding \begin{math} e_{average}\end{math} serves as \begin{math} y_i \end{math}'s token embedding in the GreenPLM. If \begin{math} y_i \end{math} repeats in the lexicon, as is frequently the case in one-to-many translations across languages, the GreenPLM framework accumulates several instances of \begin{math} e_{average} \end{math} and in turn averages them at the end of the procedure to generate a final token embedding. 

Our approach leaves the original token embeddings of source language words, tokenized word pieces, and special characters in \begin{math} S_{vocab}\end{math} untouched in the translated PLM; instead, we merely expand the vocabulary size by adding unseen target language words in the bilingual lexicon. For example in Spanish, the GreenPLM has a vocabulary size of 119,999 words, while the corresponding Spanish monolingual BERT only has 31,002. 

\subsection{Continued Pre-training} 
Continued pre-training has long been validated as one of the most effective strategies to enhance PLMs, with several possible strategies including masked language modeling (MLM) \cite{devlin-etal-2019-bert} and translation language modeling (TLM) \cite{https://doi.org/10.48550/arxiv.1901.07291}. In addition, some special techniques, such as contrastive learning, can be combined with pre-training tasks. Here we use MLM as the basic pre-training task in our continued pre-training experiments. To generate a GreenPLM+ model for each language, the translated PLM is further pre-trained with a batch size of 1024 and a sequence length of 128 symbols by utilizing 10\% of the OSCAR dataset \cite{OrtizSuarezSagotRomary2019}. Each model was pre-trained for 5 epochs and every 10,000 pre-training steps were saved.

The experiments presented here examine GreenPLM+ models in seven languages: Spanish, Indonesian, Tagalog, Turkish, Italian, Romanian, and Persian because they are the only languages with a moderate amount of language resources. An attempt to further pre-train a GreenPLM for Maltese fails because the available resources (17 MB in the OSCAR corpus) are too scarce to maintain training stability.

\subsection{Alternatives to LWM}
We found that LWM worked most effectively as a method to bridge the embedding spaces between models, and therefore present experimental results using that method. However, we also tried several other logical alternatives that were not as effective. We describe those methods here for completeness and report detailed results of these methods in Spanish, Turkish, and Tagalog in the supplementary material. Because these methods are strictly worse, we did not further experiment on other languages.

\textbf{Vocabulary Expansion (VE)} refers to expanding the vocabulary based on the vocabulary of the source language PLM. Given \textbf{a bilingual lexicon} \emph{lexicon} =\begin{math}\{x_i, y_i\} \end{math} and the PLM \emph{S} of the source language with a vocabulary of \begin{math} S_{vocab} \end{math} = \begin{math} \{w_1, w_2, ... , w_i, ... , w_n\} \end{math} and token embeddings \begin{math} S_{embedding} \end{math}, VE will walk through \begin{math} S_{vocab} \end{math} and find their corresponding embeddings by \begin{math} S_{embedding} \end{math} and \emph{lexicon}. If no matching entries exist, this strategy will keep the original tokens. If there is one \begin{math} y_i \end{math} corresponding to multiple \begin{math} x_i \end{math}, VE will keep the first \begin{math} x_i \end{math}'s embedding as the translated embedding. Then, VE will merge all \begin{math} W_i \end{math} as the vocabulary of the new model, where translated embeddings are used for the newly extended vocabulary.

\textbf{Vocabulary One-on-one Mapping (VOM)} adopts a similar methodology to VE but only uses one translation for multiple entries. While both LWM and VE increase the vocabulary sizes, introducing more parameters, VOM yields a GreenPLM with the same vocabulary sizes.

\textbf{Vocabulary Translation Mapping (VTM)} is the most basic strategy being tested. It cuts sentences according to space delimitation and uses bilingual lexicons to directly transfer the tokens into another language. Then, the newly generated sentences are directly fed to the source language PLM for downstream tasks.

\section{Materials and Evaluation}
\subsection{Language and Model Choices}\label{subsec3}
English was chosen as a source language since it has the largest pre-training corpus size and the largest body of existing PLMs available. For target languages, we included 18 languages from five language families (Table~\ref{tab:1}). According to \cite{joshi-etal-2020-state}, the chosen languages fall under various language resources' levels ranging from the extremely low-resource level ``1 - The Scraping-Bys" to the high-resource level ``5 - The Winners". 

\begin{table}[]
\centering
\resizebox{\columnwidth}{!}{%
\begin{tabular}{cccc}
\toprule
\textbf{Language}     & \textbf{Language Family} & \textbf{Language Resource} & \textbf{Monolingual PLMs' Corpus Size} \\ \midrule
Spanish       & Indo-European     & 5 - Winners                                                           & 2996.01 million words         \\
Indonesian    & Austronesian      & 3 - Rising Stars                                                      & 220 million words             \\
Turkish       & Turkic            & 4 - Underdogs                                                         & 440.50 million words          \\
Tagalog       & Austronesian      & 3 - Rising Stars                                                      & 39 million words              \\
Italian       & Indo-European     & 4 - Underdogs                                                         & 2050 million words            \\
Romanian      & Indo-European     & 3 - Rising Stars                                                      & 2421.3 million words          \\
Persian       & Indo-European     & 4 - Underdogs                                                         & 1300 million words            \\
Maltese       & Afro-Asiatic      & 2 -  Hopefuls                                                         & -                             \\
Māori         & Austronesian      & 1 - Scraping-Bys                                                      & -                             \\
Wolof         & Niger–Congo       & 2 - Hopefuls                                                          & -                             \\
Luganda       & Niger–Congo       & 1 - Scraping-Bys                                                      & -                             \\
Ilokano       & Austronesian      & 1 - Scraping-Bys                                                      & -                             \\
Hausa         & Afro-Asiatic      & 2 - Hopefuls                                                          & -                             \\
Bulgarian     & Indo-European     & 3 - Rising Stars                                                      & -                             \\
Latvian       & Indo-European     & 3 - Rising Stars                                                      & -                             \\
Ancient Greek & Indo-European     & -                                                                     & -                             \\
Norwegian     & Indo-European     & 1 - Scraping-Bys                                                      & -                             \\
Danish        & Indo-European     & 3 - Rising Stars                                                      & -                             \\ \bottomrule
\end{tabular}%
}
\caption{Languages represented in this work, representing five language families and a variety of settings with regard to resource availability.}
\label{tab:1}
\end{table}

In these experiments, we used the \emph{bert-base-uncased} English model as the source PLM, since it has the broadest coverage in academic and industrial usage. Since the model is relatively lightweight it also allows for extensive experiments on various NLP tasks across languages, even with limited computational resources. Nevertheless, the GreenPLM framework is independent of the particular structure of PLMs beyond the embedding layer, and hence can theoretically be applied to larger models to benefit from continuing advances in large-scale PLMs. 

\subsection{Experimental Settings}\label{subsec3}
\subsubsection{Bilingual Lexicons}
This study uses bilingual lexicons from four sources. For relatively high-resource languages, our lexicons are from MUSE \cite{conneau2017word}, PanLex \cite{kamholz-etal-2014-panlex}, and OPUS \cite{tiedemann-nygaard-2004-opus}. For Maltese, Māori, Wolof, Luganda, Ilokano, and Hausa, we used lexicons released from \cite{https://doi.org/10.48550/arxiv.2203.09435} directly. Detailed statistics of the lexicons could be accessed in the supplementary material.

\subsubsection{Evaluation}
Most chosen languages have available benchmark datasets for performance comparison. For example, \cite{CaneteCFP2020} as the Spanish benchmark; IndoLEM in \cite{koto-etal-2020-indolem} for Indonesian; \cite{stefan_schweter_2020_3770924}, \cite{cruz2019evaluating,cruz2020establishing,cruz2020investigating}, \cite{stefan_schweter_2020_4263142}, \cite{dumitrescu-etal-2020-birth} and \cite{Farahani_2021} for Turkish, Tagalog, Italian, Romanian, and Persian, respectively. For the 11 low-resource languages with only NER and POS tasks, GreenPLM was tested on MasakhaNER \cite{adelani-etal-2021-masakhaner} and universal dependencies \cite{10.1162/coli_a_00402}. For benchmark datasets with multiple tasks, we divided the tasks into three main categories: POS, NER, and text classification for a by-category analysis of our GreenPLM framework. Specifically, for continued pre-training, this study uses \emph{batch size} \begin{math} \times \end{math} \emph{sequence length} \begin{math}
\times \end{math} \emph{training steps} 
as a metric to evaluate pre-training efforts.

Our translated PLMs were evaluated by performance comparison with existing models on various NLP tasks, including part-of-speech tagging (POS), named entity recognition (NER), sentiment analysis, sentence classification, and paraphrase identification. We used the F1 score for the NER task, accuracy in POS and most classification tasks, Pearson's coefficient in the tweet ordering task of Indonesian, and hamming loss in the Dengue task of Tagalog. Specifically, We re-scaled rare metrics like hamming loss and Pearson's coefficient to fit a 0 to 100 scale and took the average of model performance for a direct comparison.

\subsubsection{Baseline Methods}
To compare the performance of GreenPLM with existing models, four deep learning models were chosen as baselines, and fine-tuned towards each evaluation task: (1) Scratch: a randomly initialized BERT model; (2) Pre-BERT SOTA: state-of-the-art models before BERT, i.e. long short-term memory (LSTM) and convolutional neural networks (CNNs) with pre-trained word embeddings, which we implement using NCRF++ \cite{yang-zhang-2018-ncrf}; (3) mBERT: multilingual BERT \cite{devlin-etal-2019-bert}; and (4) monolingual BERT models, where available. 
In addition, GreenPLM+, the advanced version of GreenPLM with a continued pre-training on a very limited dataset, was also compared with the above models. Only uncased models were tested in this study because most bilingual lexicons are uncased.

\section{Results}\label{sec4}
Our experiments test baseline, GreenPLM, and GreenPLM+ models on 18 languages across various categories of tasks listed in Table~\ref{tab:results}. Detailed results on each specific task could be accessed in the \textit{supplementary material}.

\begin{table*}[]
\centering
\resizebox{\textwidth}{!}{%
\begin{tabular}{cccccccc}
\toprule
  \textbf{Language}            & \textbf{Task by Category}            & \textbf{Scratch} & \textbf{Pre-BERT SOTA} & \textbf{mBERT} & \textbf{monolingual BERT} & \textbf{GreenPLM} & \textbf{GreenPLM+}                                                                         \\ \midrule
Spanish       & POS, NER, Classification*3   & 67.38                             & 81.68                                   & 88.04                           & \textbf{89.37}            & 85.27                              & 88.08 (0.7\%)                                                                                               \\
Indonesian    & POS, NER*2, Classification*3 & 63.67                             & 75.17                                   & 81.09                           & 84.56                                      & 81.08                              & \textbf{84.57} (1.4\%)                                                                    \\
Turkish       & POS*2, NER                   & 70.38                             & 90.41                                   & 92.97                           & 93.07                                      & 90.70                              & \textbf{93.09} (2.5\%)                                                                     \\
Tagalog       & Classification*3             & 77.41                             & 77.44                                   & 89.43                           & 85.77                                      & \textbf{89.59}    & \textbf{89.59} (0\%) \\
Italian       & POS*2                        & 83.47                             & 96.39                                   & 97.44                           & 97.38                                      & 95.98                              & \textbf{97.63} (0.5\%)                                                                     \\
Romanian      & POS, NER, Classification*2   & 67.08                             & 82.26                                   & 86.5                            & 88.72                                      & 85.39                              & \textbf{88.79} (1.2\%)                                                                     \\
Persian       & NER*2, Classification*3      & 84.9                              & 90.52                                   & 91.96                           & 93.59                                      & 91.05                              & \textbf{93.60} (0.8\%)                                                                     \\
Maltese       & POS, NER                     & 40.47                             & 70.79                                   & 73.30                           & -                                          & \textbf{76.93}    & -                                                                                                           \\
Māori         & NER                          & 53.33                             & \textbf{89.44}         & 81.97                           & -                                          & 79.21                              & -                                                                                                           \\
Wolof         & POS, NER                     & 48.45                             & \textbf{72.99}         & 72.72                           & -                                          & 72.71                              & -                                                                                                           \\
Luganda       & NER                          & 30.39                             & \textbf{74.65}         & 74.19                           & -                                          & 74.48                              & -                                                                                                           \\
Ilokano       & NER                          & 29.65                             & 64.58                                   & 67.44                           & -                                          & \textbf{67.98}    & -                                                                                                           \\
Hausa         & NER                          & 50.66                             & 82.38                                   & \textbf{85.58} & -                                          & 84.97                              & -                                                                                                           \\
Bulgarian     & POS                          & 92.16                             & 97.93                                   & \textbf{99.06} & -                                          & 97.96                              & -                                                                                                           \\
Latvian       & POS                          & 89.53                             & 94.48                                   & \textbf{96.33} & -                                          & 94.59                              & -                                                                                                           \\
Ancient Greek & POS                          & 89.42                             & 95.00                                   & \textbf{96.00} & -                                          & 95.08                              & -                                                                                                           \\
Norwegian     & POS                          & 92.82                             & 96.50                                   & \textbf{97.61} & -                                          & 96.75                              & -                                                                                                           \\
Danish        & POS                          & 82.37                             & 94.50                                   & \textbf{97.80} & -                                          & 95.90                              & -                                                                                                           \\ \bottomrule
\end{tabular}%
}
\caption{Tasks and experimental results in six settings. Numbers represent average performance across language-specific evaluation tasks; since the available tasks are different for each language, numbers are only comparable within their own row. Parentheticals in the last column present the continued pre-training efforts required to reach performance equal to the corresponding monolingual model compared to training from scratch; for Tagalog, this number is 0\% because we didn't continue pretraining GreenPLM as it has outperformed the monolingual one.} 
\label{tab:results}
\end{table*}

\subsection{Base GreenPLM}
In the base setting where models are translated using only LWM, we observe that the GreenPLM performs much better than random BERT in all cases, and better than Pre-BERT SOTA models in 15 out of 18 languages, with the exception of sequence labeling tasks in 3 languages. This suggests that indeed, the bilingual lexicon based approach we take with GreenPLM is not merely successful due to task-specific fine-tuning, but rather achieves performance gains due to cross-linguistic knowledge transfer from the source PLM. Though GreenPLM generally does not surpass mBERT in performance, our method provides at least comparable performance on the examined languages at nearly no cost, whereas mBERT itself requires large multilingual training costs.

Moreover, in some language settings our method is surprisingly effective. In \textbf{Tagalog}, the GreenPLM wins over all baselines and shows a large improvement over the monolingual BERT. This may imply that the original Tagalog BERT was not pre-trained with sufficient data. In the low-resource settings of \textbf{Maltese} and \textbf{Ilokano}, the GreenPLM also outperforms all existing baselines.


\subsection{GreenPLM+: Continued Pre-training}
As a further enhancement to the GreenPLM, we applied a lightweight continued pre-training step with very limited training cost and evaluated the results in the seven languages mentioned above with access to sufficiently-sized pre-training corpora. In the last column of Table \ref{tab:results}, we evaluate GreenPLM+ models at the first point at which they outperform mBERT or monolingual BERT for the first time during the continued pre-training, and note in parens the percentage of pre-training efforts required to reach this level of performance
compared to pre-training a PLM from scratch. We observe that GreenPLM+ outperforms all monolingual BERT at a relatively low cost, except in Spanish. For example, GreenPLM+ outperforms the original PLM with only \textbf{0.68\%} of the pre-training efforts
for Indonesian. For Spanish, GreenPLM+ outperforms multilingual BERT and achieves comparable performance over the monolingual BERT when given \textbf{0.7\%} of the pre-training efforts. 
We further compared continuously pre-training GreenPLM and mBERT under the same settings in Indonesian, Romanian, and Persian. GreenPLM could catch up and outperform mBERT given minimal costs. In Figure~\ref{fig:persian}(b), we show a visualized comparison of Persian.

\section{Discussion}
\begin{figure*}[]
  \centering
  \includegraphics[width=\textwidth]{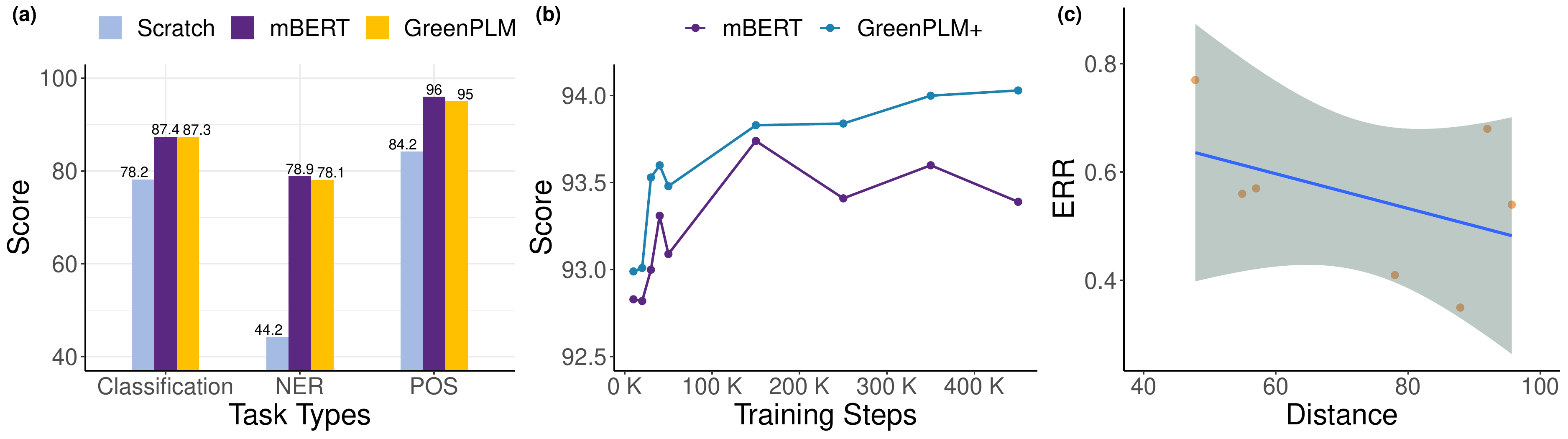}
  \caption{Performance comparison of GreenPLM under different settings. \textbf{(a)} Results across NER, POS, and Classification task types. \textbf{(b)} Comparison between continued pre-training GreenPLM+ versus multilingual BERT in Persian. \textbf{(c)} We observe a linear relationship between language distance and error reduction rate.}
\label{fig:persian}
\end{figure*}

\paragraph{Impact on Sustainable Development.} High computational requirements for pre-training LMs or even pre-BERT deep learning models have been a major obstacle for researchers, where a slight performance boost may come at the expense of dramatically increased model complexity. Our approach generates translated PLMs for medium-level and low-level resource languages that successfully outperform pre-BERT SOTA deep learning models at extremely low pre-training costs, both in terms of economic costs and CO\begin{math}_{2}\end{math} emissions. In contrast to prior approaches for low-cost training of PLMs \cite{https://doi.org/10.48550/arxiv.2111.04130}, the GreenPLM framework is task-agnostic 
All computation for PLM translation in the basic GreenPLM approach can be executed on a personal computer within 10 seconds. For continued pre-training with the GreenPLM+ model, we demonstrate comparable results to a fully pre-trained monolingual model using less than 10\% of the dataset size in less than 8 hours on an 8x RTX3090 system. 

\paragraph{Impacts on Language Equality.} Unequal language resources have limited recent advances in PLMs from being equally shared by speakers of all the world's languages. While developing larger corpora for pre-training is not always feasible, here we leverage a more widely available language resource -- bilingual lexicons -- to translate PLMs from higher- to lower-resource settings.
This framework can serve as a basic starting point to transfer models across languages and reduce both language-based and economic disparities in the availability of PLMs in different linguistic and socioeconomic backgrounds. 

Experimental results show that GreenPLM+ performs better than existing monolingual and multilingual models in languages with small but still substantial pre-training corpus sizes, such as Indonesian, Turkish, and Tagalog. These languages are rated as "4 - Underdogs" or "3 - Rising Stars" in the \cite{joshi-etal-2020-state} classification of resource availability. We note with an example from Persian in Figure~\ref{fig:persian} (b) that in these settings, the base GreenPLM model commonly starts off with lower performance than mBERT, but the GreenPLM+ quickly catches up after continued pre-training with the same number of steps.

For languages with fewer resources in the `2 - Hopefuls' and `1 - Scraping-Bys' categories by \cite{joshi-etal-2020-state}, where substantial pre-training corpora are not available for continued pre-training, we find that mBERT retains better performance than the base GreenPLM. Nevertheless, the simple and nearly zero-cost transfer method presented here achieves performance generally surpassing Pre-BERT SOTA models, and comparable to mBERT, suggesting that base GreenPLM models can be productively used to transfer PLMs to languages where only bilingual lexicons exist. 


\paragraph{Applicability for Downstream Tasks.} The evaluations in this work examined downstream tasks in the categories of POS tagging, NER, and classification tasks, corresponding to word-level, span-level, and sentence-level understanding. Results in Figure~\ref{fig:persian}(a) show that the GreenPLM, compared to random BERT models, has the most remarkable improvement on POS tasks, followed by NER and classification tasks. However, compared to mBERT, GreenPLM falls short in POS, followed by NER and classification tasks. This suggests that GreenPLM transfers sentence-level understanding the best in downstream tasks, possibly because embedding-based strategies of our framework focus mainly on semantic information and lack the ability to capture the variability of syntactic information in natural languages.

\paragraph{Language Distance.} In addition, we use a language distance calculator from eLinguistics \cite{article} as a reference and quantify the performance by the error reduction rate (ERR) metric because languages are classified according to their diachronic relatedness in linguistic typology. Corresponding results in Figure~\ref{fig:persian}(c) show that the closer the selected language is to English, the better its GreenPLM will perform. This is not surprising since GreenPLM models retain the parameters from the original English BERT being translated; nevertheless, it suggests that linguistic distance is likely to be a crucial consideration in cross-lingual knowledge transfer, supporting existing findings \cite{lin-etal-2019-choosing}.

\paragraph{Future Work.} Though this paper uses English as a source language and starts from an English PLM to generate target language PLMs, a similar approach could be used to merge PLMs from multiple languages for cross-enhancement to generate a single target language PLM. In addition, GreenPLM could be extended to multi-modal and multi-domain settings where bilingual dictionaries could facilitate information exchange; for example, could a similar approach help integrate biomedical information using a targeted lexicon? Finally, it remains to investigate how the GreenPLM approach performs with more state-of-the-art BERT-class PLMs such as RoBERTa, as well as with PLMs of different architectures such as BART \cite {lewis-etal-2020-bart}.

\paragraph{Limitations.} This work has two main limitations. First, it only covers uncased PLMs at the current stage because most bilingual lexicons are uncased, and capitalization could affect model performance. 
Second, there is no mechanism to rank the importance of single entries in bilingual lexicons. Therefore, newly generated models generally have a more extensive vocabulary than the original PLM.

\section{Conclusion}
This paper proposes a bilingual lexicon-based framework GreenPLM that ``translates'' a monolingual PLM to other languages with no additional cost. We verify its effectiveness in 18 languages and further pre-train the resulting models to validate their performance when given minimal pre-training costs. Our results provide evidence that the GreenPLM+ (GreenPLM with very limited continued pre-training) outperforms monolingual BERT at much lower computational costs than current frameworks, such as the continued pre-training of mBERT. This framework is easy to incorporate into existing PLMs and can be adapted to various languages given the relatively wide availability of bilingual lexicons. We hope that this work can make a positive impact on promoting green AI and reducing language disparities in access to contemporary NLP technologies.

\section*{Acknowledgments}
This research was supported by the National Natural Science Foundation of China (No.
62206243).

\small
\bibliographystyle{named}
\bibliography{ijcai23}

\end{document}